\documentclass[sigconf]{acmart}

\usepackage{amsmath}
\usepackage{algorithm}
\usepackage{enumitem}
\usepackage{picinpar}

\usepackage{graphicx}

\usepackage{algorithmicx}
\usepackage[noend]{algpseudocode}
\usepackage{subfigure}
\usepackage{multirow}
\usepackage{color}
\usepackage{balance}
\usepackage{enumitem}
\usepackage{hhline}
\usepackage[normalem]{ulem}
\usepackage{booktabs}
\usepackage{wrapfig}
\usepackage{cancel}
\usepackage{hyperref}
\usepackage{makecell}

\newcommand{\bL}{\ensuremath{\mathcal{L}}}

\newcommand{\bG}{\ensuremath{\mathcal{G}}}
\newcommand{\bN}{\ensuremath{\mathcal{N}}}

\newcommand{\bT}{\ensuremath{\mathcal{T}}}

\newcommand{\bP}{\ensuremath{\mathcal{P}}}
\newcommand{\bD}{\ensuremath{\mathcal{D}}}
\newcommand{\bO}{\ensuremath{\mathcal{O}}}

\renewcommand{\vec}[1]{\ensuremath{\mathbf{#1}}}

\newcommand{\stitle}[1]{\vspace{0.8mm} \noindent {\bf #1}}
\newcommand{\cp}{specific prompts{}}
\newcommand{\op}{holistic prompts{}}

\newcommand{\eg}{{\it e.g.}}

\newcommand{\ie}{{\it i.e.}}
\newcommand{\etc}{{\it etc.}}

\newcommand{\method}[1]{\textsc{#1}}
\newcommand{\model}{\method{SAMGPT}{}}

\newcommand{\eat}[1]{}

\newcommand{\stkout}[1]{\ifmmode\text{\sout{\ensuremath{#1}}}\else\sout{#1}\fi}

\copyrightyear{2025}
\acmYear{2025}
\setcopyright{acmlicensed}
\acmConference[WWW '25] {Proceedings of the ACM Web Conference 2025}{April 28--May 2, 2025}{Sydney, NSW, Australia.}
\acmBooktitle{Proceedings of the ACM Web Conference 2025 (WWW '25), April 28--May 2, 2025, Sydney, NSW, Australia}
\acmISBN{979-8-4007-1274-6/25/04}
\acmDOI{10.1145/3696410.3714828}%%
\AtBeginDocument{%
  }

\begin{document}

\title{SAMGPT: Text-free Graph Foundation Model for Multi-domain Pre-training and Cross-domain Adaptation}

\author{Xingtong Yu}
\affiliation{%
 \institution{Singapore Management University}
 \city{Singapore}
  \country{Singapore}}
\email{xingtongyu@smu.edu.sg}

\author{Zechuan Gong}
\affiliation{%
 \institution{University of Science and Technology of China}
 \city{Hefei}
  \country{China}}
\email{gongzechuan@mail.ustc.edu.cn}

\author{Chang Zhou}
\affiliation{%
 \institution{University of Science and Technology of China}
 \city{Hefei}
  \country{China}}
\email{zhouchang21sy@mail.ustc.edu.cn}

\author{Yuan Fang$^{*}$}
\affiliation{%
  \institution{Singapore Management University}
  \city{Singapore}
  \country{Singapore}}
\email{yfang@smu.edu.sg}

\author{Hui Zhang$^*$}
\affiliation{%
  \institution{University of Science and Technology of China}
  \city{Hefei}
  \country{China}}
\email{fzhh@ustc.edu.cn}

\thanks{
    $^*$ Corresponding authors.
}

\renewcommand{\shortauthors}{Xingtong Yu, Zechuan Gong, Chang Zhou, Yuan Fang, and Hui Zhang}

\begin{abstract}
Graphs are able to model interconnected entities in many online services, supporting a wide range of applications on the Web. This raises an important question: How can we train a graph foundational model on multiple source domains and adapt to an unseen target domain? A major obstacle is that graphs from different domains often exhibit divergent characteristics. Some studies leverage large language models to align multiple domains based on textual descriptions associated with the graphs, limiting their applicability to text-attributed graphs. For text-free graphs, a few recent works attempt to align different feature distributions across domains, while generally  neglecting structural differences. In this work, we propose a novel Structure Alignment framework for text-free Multi-domain Graph Pre-Training and cross-domain adaptation (\model). It is designed to learn multi-domain knowledge from graphs originating in multiple source domains, which can then be adapted to address applications in an unseen target domain. Specifically, we introduce a set of \emph{structure tokens} to harmonize structure-based aggregation across source domains during the pre-training phase. Next, for cross-domain adaptation, we design dual prompts, namely, \emph{\op} and \emph{\cp}, which adapt unified multi-domain structural knowledge and fine-grained, domain-specific information, respectively, to a target domain. Finally, we conduct comprehensive experiments on seven public datasets to evaluate and analyze the effectiveness of \model.
\end{abstract}





\maketitle

\section{Introduction}\label{sec.intro}
How to build foundation models has emerged as an important question, paving a plausible path toward artificial general intelligence. In natural language processing, recent works \cite{touvron2023llama,achiam2023gpt} have demonstrated the capabilities of universal foundation models. They are trained on a wide variety of data from multiple domains, and can be further adapted to solve a diverse range of tasks. Other than natural languages, the World Wide Web has become a vast knowledge repository, connecting an enormous amount of entities to form extensive and complex graphs. These graphs enable diverse Web applications, including social network analysis~\cite{miyake2024netevolve,jiang2024ragraph}, Web mining~\cite{fangjf2023exgc,agarwal2022graphnli}, and recommendation systems~\cite{ma2024temporal,ni2024graph}. 
Given the rich graph data on the Web, can we build a universal graph model based on multi-domain graphs, to address various downstream graph-centric applications \cite{liu2023towards}? 

Traditional supervised graph learning struggles to build universal models. 
These approaches require retraining a new graph neural network (GNN) \cite{kipf2016semi,fangjf2023eva,yu2023learning} or graph transformer \cite{yun2019graph,rampavsek2022recipe,ying2021transformers} for each new task, relying on abundant task-specific labeled data. 
In contrast, more recent graph pre-training methods \cite{hu2020gpt,qiu2020gcc,velivckovic2018deep} attempt to learn universal properties from unlabeled graphs in a self-supervised manner, which can be subsequently adapted to a downstream task with some task-specific labels through fine-tuning \cite{kipf2016variational,velivckovic2018deep,qiu2020gcc} or prompt learning  \cite{liu2023graphprompt,sun2023all,yu2025node}.
However, in most existing graph pre-training approaches, the pre-training and downstream graphs originate from the same dataset \cite{velivckovic2018deep,you2020graph,liu2023graphprompt,sun2023all}, a practice we refer to as \emph{single-domain} methods, which fall short of building a universal, \emph{multi-domain} graph model from multiple graph datasets.

\stitle{Research problem.}
Thus, it is crucial to pre-train a graph model on a wide range of multi-domain (\ie, multi-dataset) graphs and achieve cross-domain adaptation. However, graph structures from different datasets often exhibit markedly distinct characteristics. For instance, the structural patterns in a social network might not be directly applicable to a citation or e-commerce graph. 
Such diversity poses significant challenges in integrating graphs from multiple domains and adapting prior knowledge to different domains. Although some studies have explored cross-domain adaptation from a single source domain~\cite{ding2021cross,hassani2022cross,wang2021pre,wang2023cross,yang2019cross}, they do not exploit multiple  source domains. Another line of work~\cite{liu2023one,tang2024higpt,xia2024opengraph} employs large language models to extract and utilize multi-domain knowledge based on textual descriptions associated with graphs, using text as a universal medium to bridge different domains. However, this limits their applicability to text-attributed graphs~\cite{zhaolearning,wen2023augmenting} and cannot be extended to general graphs without textual descriptions. Few recent studies~\cite{zhao2024all,yu2024text} have explored multi-domain pre-training on text-free graphs, but they focus on aligning the divergent feature spaces and homophily patterns across multi-domain graphs, while overlooking the structural differences across  domains. 

\stitle{Challenges and insights.}
In this paper, we propose \model, a graph foundation model with \textbf{S}tructural \textbf{A}lignment for text-free \textbf{M}ulti-domain \textbf{G}raph \textbf{P}re-\textbf{T}raining, to facilitate cross-domain adaptation. This is non-trivial due to two key challenges.

\begin{figure}[t]
\centering
\includegraphics[width=1\linewidth]{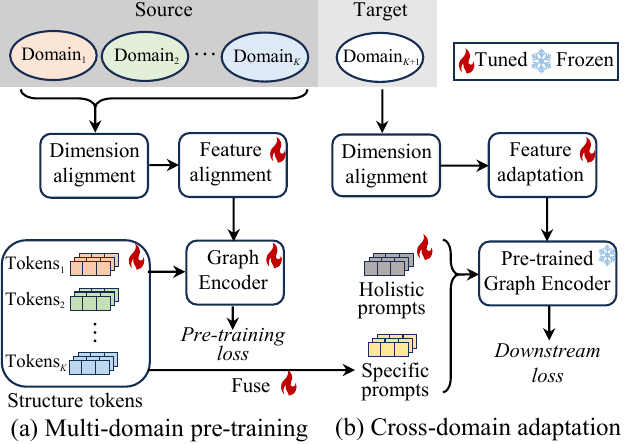}%
\vspace{-1mm}%
\caption{Motivation of \model.}
\label{fig.intro-motivation}
\end{figure}

First, \textit{how do we harmonize structural variance across multiple domains during pre-training?} 
Graphs from different domains often exhibit distinct structural and topological characteristics, such as average node degree, shortest path length and clustering coefficient, as depicted in Table~\ref{table.datasets}. %~\ref{app.dataset}.
Thus, merging multi-domain graphs without structure alignment during pre-training may cause interference rather than synergy, leading to suboptimal performance.
In \model, we propose \textit{structure tokens} to align structural distributions across multiple domains, as shown in Fig.~\ref{fig.intro-motivation}(a). Each domain is equipped with a series of structure tokens, which modify the structure-based aggregation in each layer of the graph encoder. These tokens are learnable vectors that capture domain-specific structural patterns, enabling the model to accommodate the unique structural characteristics of each domain during pre-training.

Second, \textit{how do we adapt multi-domain prior  structural knowledge to cross-domain downstream tasks?} 
Multi-domain prior knowledge includes not only holistic knowledge across source domains, but also domain-specific knowledge from each domain. Therefore, in \model, we %extend beyond feature alignment by proposing 
propose dual \textit{structural prompts}, comprising a set of \textit{\op} and a set of \textit{\cp}, thus facilitating the adaptation of both holistic and domain-specific knowledge to downstream tasks, as illustrated in Fig.~\ref{fig.intro-motivation}(b). On one hand, the \op\ consist of learnable vectors that holistically align the target domain's structural characteristics with the unified pre-trained knowledge from all source domains. On the other hand, \cp\ integrate multi-domain structure tokens in a learnable mixture to align the target domain with  knowledge from each source domain, capturing domain-specific structural information for finer-grained adaptation.

\stitle{Contributions.}
In summary, we make the following contributions in this work.
(1) We propose \model, a text-free graph foundation model with structure alignment for multi-domain graph pre-training and cross-domain adaptation.
(2) For pre-training, we propose structure tokens to align structural distributions across domains, training a universal foundation model with multi-domain graphs.
(3) For downstream adaptation, we propose a dual-prompt strategy, using \op\ to leverage holistic prior structural knowledge and \cp\ to facilitate finer-grained, domain-specific structural adaptation.
(4) We conduct extensive experiments on seven benchmark datasets, demonstrating the superior performance of \model\ compared to state-of-the-art methods.

\section{Related Work}

We review related literature on pre-training, cross-domain transfer learning, and multi-domain pre-training for graph data.

\stitle{Graph pre-training.}
Graph pre-training methods aim to extract inherent properties of graphs, often utilizing self-supervised learning approaches, which can be either generative \cite{hu2020gpt,li2023s,hou2022graphmae,jiang2023incomplete} or contrastive \cite{velivckovic2018deep,xia2022simgrace,xu2021self,li2022mining}. The pre-trained model is then employed to address downstream tasks through fine-tuning \cite{you2020graph,velivckovic2018deep,qiu2020gcc} or parameter-efficient adaptation methods, notably prompt-based learning \cite{sun2022gppt,liu2023graphprompt,yu2023generalized,fang2022universal}. However, these methods typically assume that the pre-training and downstream graphs originate from the same domain, such as different subgraphs of a large graph \cite{you2020graph,yu2023hgprompt} or collections of similar graphs within the same dataset \cite{hu2020gpt,qiu2020gcc}, failing to account for multiple domains in either pre-training or downstream graphs.

\stitle{Graph cross-domain transfer.}
This line of work aims to transfer single-source domain knowledge to a different target domain by leveraging domain-invariant properties across domains~\cite{ding2021cross,hassani2022cross,wang2021pre,wang2023cross}. However, they rely exclusively on a single source domain, failing to harness the extensive knowledge available across multiple domains. Additionally, these approaches are often tailored to specific tasks or domains \cite{ding2021cross,hassani2022cross,wang2021pre,wang2023cross}, limiting their generalization.

\stitle{Multi-domain graph pre-training.}
In the context of graphs from multiple domains, recent works \cite{liu2023one,tang2024higpt,xia2024opengraph} utilize large language models to align node features from different domains through textual descriptions, thereby limiting their applicability to text-attributed graphs \cite{zhaolearning,wen2023prompt,zhang2024text}. For graphs without textual attributes, GraphControl \cite{zhu2024graphcontrol} applies ControlNet~\cite{zhang2023adding} to incorporate target domain node features with the pre-trained model, while neglecting the alignment among multiple source domains. Another recent study proposes GCOPE~\cite{zhao2024all}, which employs domain-specific virtual nodes that interconnect nodes across domains, facilitating the alignment of feature distribution and homophily patterns. Meanwhile, MDGPT~\cite{yu2024text} pre-trains domain-specific tokens to align feature semantics across various domains. However, these studies do not account for structural variance across different domains, hindering their effectiveness in integrating multi-domain knowledge. On a related front, multi-task pre-training techniques \cite{wang2022multi,yu2023multigprompt} employ pretext tokens for each pre-training task. It is important to note that they address a distinct problem, aiming to overcome potential interference among multiple tasks within a single domain, rather than interference across multiple domains. 
%In our work, we propose structure tokens and dual prompts to overcome the limitations of current multi-domain graph pre-training methods. 
%Note that multi-task pre-training aims to reduce interference among different pre-training tasks within a single domain, distinct from our objective of multi-domain pre-training.

\section{Preliminaries}
In this section, we provide technical background, and outline the scope of our work.

\stitle{Graph encoder.}
A \emph{graph} is defined as \( G = (V, E, \vec{X}) \), where \( V \) is the set of nodes, \( E \) is the set of edges, and \( \vec{X} \in \mathbb{R}^{|V| \times d} \) is the node feature matrix with each row \( \vec{x}_i \) representing the feature vector of node \( v_i \in V \). A collection of graphs is denoted as \( \mathcal{G} \).

\begin{figure*}[t]
\centering
\includegraphics[width=1\linewidth]{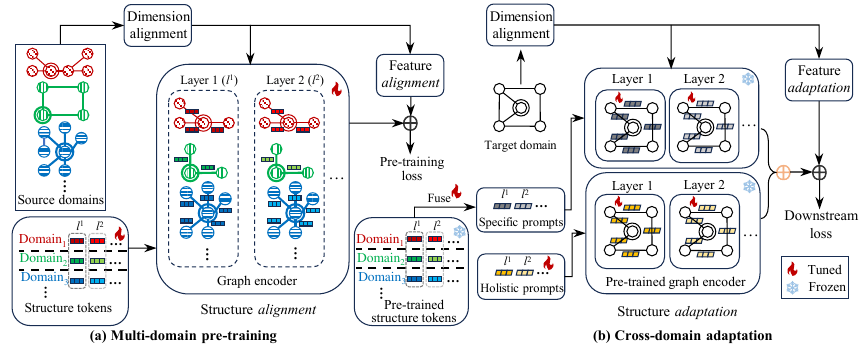}%
\vspace{-1mm}%
\caption{Overall framework of \model.}
\label{fig.framework}
\end{figure*}

%\stitle{Graph encoder.}
Message-passing GNNs are a common choice for encoding graph representations \cite{wu2020comprehensive}. Specifically, each node updates its embedding by receiving and aggregating features or embeddings from its neighbors. By stacking such message-passing layers, information can propagate recursively throughout the graph. Therefore, the node embeddings are encoded based on both input features and graph structure. Let us denote the embedding of node \( v \) at the \( l \)-th layer as \( \vec{h}^l_v \), which is derived from the features or embeddings in the preceding layer as follows.
\begin{align}
    \vec{h}^l_v = \mathtt{Aggr}(\vec{h}^{l-1}_v, \{\vec{h}^{l-1}_u : u\in\bN_v\}; \theta^l),
\end{align}
where \( \mathcal{N}_v \) denotes the set of neighboring nodes of \( v \), \( \theta^l \) represents the learnable parameters in layer \( l \), and \( \mathtt{Aggr}(\cdot) \) stands for the neighborhood aggregation function. In the first layer, the node embedding \( \vec{h}^0_v \) is initialized as the input feature vector $\vec{x}_v$. We denote the output node embedding after the last layer as \( \vec{h}_v \), which is a row in the node embedding matrix \( \vec{H} \). Overall, the multi-layer message-passing process can be abstracted as a \emph{graph encoder}, as follows. 
\begin{align}
    \vec{H} = \mathtt{GE}(G,\vec{X};\Theta),
\end{align}
where $\mathtt{GE}$ denotes a graph encoder, \( \Theta = \{\theta^1, \theta^2, \ldots\} \) is the full set of trainable parameters for the graph encoder.

\stitle{Multi-domain pre-training with feature alignment.}\label{sec.pre.multi-domain}
Consider a set of unlabeled graphs \( \mathcal{G}_S = \{ G_1, G_2, \ldots, G_K \} \) for pre-training, where each graph \( G_i \) belongs to a specific \emph{source domain} \( D_{S_i} \in \mathcal{D}_S \). 
%Each domain \( D_{S_i} \) is characterized by its structure (node\& edge), and feature distributions, represented respectively as \( P^{(V)}_{S_i} \), \( P^{(E)}_{S_i} \), and \( P^{(\vec{X})}_{S_i} \). 
Thus, we have graph-domain pairs \( \{(G_i, D_{S_i}): i \in \{ 1, 2, \ldots, K\}\} \). 
%Note that some graphs may belong to the same domain, meaning there could be $D_{S_i} = D_{S_j}$ for some $i\ne j$.
%\( \exists i \neq j, D_{S_i} = D_{S_j} \).
%Subsequently, the input to the pre-training phase consists of the unlabeled source domain graphs $\mathcal{G}_S$. 

As different domains exhibit distinct feature distributions, previous works \cite{zhao2024all,yu2024text} have proposed solutions to align feature dimensions and semantics, which can be directly employed in our work. 
Given a graph \( G_{i} = (V_{i}, E_{i}, \vec{X}_{i}) \) from the source domain \( D_{S_i} \), we first align the dimensions of its feature matrix:
\begin{align}\label{eq.pre-train.dim-align}
    \tilde{\Vec{X}}_{i} = \mathtt{DAL}_{S_i}(\Vec{X}_{i}),
\end{align}
where \( \mathtt{DAL}_{S_i} \colon \mathbb{R}^{|V| \times d_{S_i}} \rightarrow \mathbb{R}^{|V| \times \tilde{d}} \) is the dimension alignment function for domain \( D_{S_i} \), transforming the original dimension $d_{S_i}$ to a common dimension $\tilde{d}$ across domains. We implement $\mathtt{DAL}$ as singular value decomposition \cite{stewart1993early} following prior art \cite{zhao2024all,yu2024text}.
Next, given the source-domain graphs $\mathcal{G}_S$ with their dimension-aligned features $\mathcal{\tilde{X}}_S=\{\tilde{\Vec{X}}_{i}: G_{i} \in \bG_{S}\}$, we further align the features to unify their semantic space across various domains. Letting $\mathtt{FAL}$ denote the feature alignment procedure, we pre-train a graph encoder with feature alignment:
\begin{align}\label{eq.pre-train.feature-align}
    \Vec{H}^{\mathtt{FAL}} = \mathtt{GE}(\mathtt{FAL}(\bG_S,\tilde{\mathcal{X}_S};\Psi);\Theta),
\end{align}
where $\Psi$ denotes the learnable parameters in $\mathtt{FAL}$, and $\Vec{H}^{\mathtt{FAL}}$ is the output node embedding matrix with feature alignment. While any feature alignment model can be employed  \cite{zhao2024all,yu2024text}, we follow the work of \citet{yu2024text} due to its superior performance.

\stitle{Cross-domain task with feature adaptation.}\label{sec:problem_definition}
For each downstream task, consider a set of graphs \( \mathcal{G}_T \) belonging to a target domain \( D_T \). The task is \emph{cross-domain} if the target domain is \emph{unseen} during pre-training,
\ie, \( \forall i \; D_T \ne D_{S_i} \). Again, since the target domain may exhibit different feature characteristics from the source domains, 
previous works \cite{zhao2024all,yu2024text} have proposed feature adaptation strategies to transfer prior multi-domain knowledge to the target domain, which can be directly integrated into our work.
Specifically, we first employ the same dimension alignment method used in the pre-training phase,
transforming the feature matrix of a downstream graph $G=(V,E,\Vec{X}) \in \bG_T$ to $\tilde{\Vec{X}}=\mathtt{DAL}_T(\Vec{X})$\label{eq.target-dimension}. We then employ a feature adaptation technique $\mathtt{FAD}$ to adapt the pre-trained model to the target domain, as follow.
\begin{align}\label{eq.feature-emb}
    \Vec{H}^\mathtt{FAD} = \mathtt{GE}(\mathtt{FAD}(G,\tilde{\Vec{X}};\Gamma);\Theta_\text{pre}),
\end{align}
where $\Gamma$ denotes the learnable parameters in $\mathtt{FAD}$, and $\Theta_\text{pre}$ is the pre-trained weights in graph encoder $\mathtt{GE}$. Here we implement $\mathtt{FAD}$ following  \citet{yu2024text}, which is paired with the feature alignment method in pre-training.

\stitle{Our scope: Few-shot classification}. For the downstream applications, we aim to solve \emph{few-shot} node and graph classification tasks. For node classification, given a graph \( G = (V, E, \vec{X}) \in \mathcal{G}_T \), each node \( v \in V \) is associated with a label \( y \in Y \), where $Y$ denotes the set of node classes. For graph classification over a set of graphs \( \mathcal{G}_T \), each graph \( G \in \mathcal{G}_T \) is associated with a label \( y \in Y \), where $Y$ denotes the set of graph classes. An \( m \)-shot classification task consists of only \( m \) labeled examples per class, along with an arbitrary number of unlabeled examples for testing. 

In particular, we focus on \emph{low-shot} settings, where \( m \) is a small number (\eg, \( m \leq 5 \)), reflecting real-world scenarios where labeled data are expensive or difficult to obtain. Due to the parameter-efficient nature of prompt learning, many previous methods for prompt learning on graphs  \cite{liu2023graphprompt,yu2023generalized,sun2023all,yu2024text,zhao2024all}  also emphasize this setting. It is expected that, as more task-specific labeled data become available, conventional fine-tuning or supervised approaches may become sufficient. 
%Therefore, our evaluation is tailored to low-resource scenarios, reflecting real-world situations where labeled data are expensive to obtain. 
%Moreover, in common graph benchmark datasets, \( m > 5 \) already constitutes a significant portion of the data (see Appendix~\ref{app.why-low-shot}).

\section{Proposed Approach: \model}

In this section, we present \model, beginning with an overview and then delving into the details of multi-domain pre-training and cross-domain adaptation.

\subsection{Overall Framework}
\model\ consists of two phases: multi-domain pre-training, and cross-domain adaptation, as shown in Fig.~\ref{fig.framework}.

In the pre-training phase, as depicted in Fig.~\ref{fig.framework}(a), we first align the feature distributions from multiple source domains following previous work \cite{zhao2024all,yu2024text}. 
Next, we introduce a set of \textit{structure tokens} designed to align the structural distributions across diverse domains. These tokens are domain-specific and are integrated into each layer of the graph encoder, modifying the structure-based aggregation at each layer. 
Finally, the structure token-enhanced graph encoder is pre-trained using a self-supervised loss  \cite{liu2023graphprompt}.
%based on a universal task template \cite{liu2023graphprompt}.

In the adaptation phase, as shown in Fig.~\ref{fig.framework}(b), we first align the feature dimension of the target domain with that of the source domains. Then, we introduce \textit{dual prompts}. The first type, \emph{\op}, are learnable vectors that integrate the target domain with the holistic structural knowledge from all source domains. The second type, \emph{\cp}, comprise learnable mixtures of pre-trained structure tokens that incorporate domain-specific topological information tailored to the target domain. These prompts are applied to each layer of the graph encoder to adjust the structure-based aggregation, while keeping the pre-trained weights of the graph encoder frozen. 
%facilitating cross-domain adaptation with lightweight tuning.

\subsection{Multi-domain Graph Pre-training with Structure Alignment}

As defined in Sect.~\ref{sec.pre.multi-domain}, we are given a set of pre-training graphs from multiple source domains, $\bG_S$. As both the features and structures of these domains can exhibit divergent distributions, effective integration of these multi-domain graphs requires aligning both.
As our work focuses on structure alignment, we follow previous feature alignment methods  \cite{zhao2024all,yu2024text}, as outlined in the preliminaries.

\stitle{Structure alignment.}
Recall that in the graph encoder, node representations are updated layer-wise through a structure-based aggregation. Each layer captures different levels of structural information. For example, the first layer aggregates one-hop neighborhood information, while the second layer incorporates a broader two-hop neighborhoods. These layer-wise structural patterns may vary significantly across domains. 

Therefore, to unify the structural characteristics in multiple source domains, we introduce learnable \emph{structure tokens}. For each domain $D_{S_i}$, we inject a series of structure tokens $\bT_{S_i}=\{\Vec{t}^l_{S_i}:  l\in \{1,\ldots,L\}\}$ into the graph encoder, where $L$ denotes the number of layers. Specifically, when encoding the graph $G_i=(V_i,E_i,\tilde{\mathbf{X}}_i$) in $D_{S_i}$, we assign structure token $\Vec{t}^l_{S_i}$ to the $l$-th layer, guiding structure-based aggregation:
\begin{align}\label{eq:structure_aggregation}
    \vec{h}^l_{v} = \mathtt{Aggr}(\vec{h}^{l-1}_{v}, \{\Vec{t}^l_{S_i} \odot \vec{h}^{l-1}_{u} : u\in\bN_v\}; \theta^l),~\forall v\in V_i,
\end{align}
%\begin{align}\label{eq:structure_aggregation}
%    \vec{h}^l_{S_i,v} = \mathtt{Aggr}(\vec{h}^{l-1}_{S_i,v}, \{\Vec{t}^l_{S_i} \odot \vec{h}^{l-1}_{S_i,u} : u\in\bN_v\}; \theta^l),
%\end{align}
where $\odot$ represents element-wise multiplication. Note that the graph encoders for feature alignment and structure alignment on all graphs share the same parameters $\Theta$.
Let $\vec{H}^{\mathtt{SAL}}_{i}$ denote the structure-aligned output node embedding matrix for $G_i$ in $D_{S_i}$, following the aggregation in Eq.~\eqref{eq:structure_aggregation}.
In general, each source domain is attached with its own set of structure tokens,
which are applied to modify the aggregation on the graph in the corresponding domain. By stacking the structure-aligned output matrix across graphs in all domains, we obtain the overall structure-aligned embedding matrix,  
$\Vec{H}^\mathtt{SAL}= \mathtt{Stack}(\vec{H}^{\mathtt{SAL}}_{1},\ldots,\vec{H}^{\mathtt{SAL}}_{K})$.
%, defined as $\bT=\{\bT_{S_i}: \forall S_i \text{ s.t. } D_{S_i} \in \bD_{S}\}$. By applying these structure tokens to all source domains, we obtain the structure aligned embedding matrix  $\Vec{H}^\mathtt{SAL}_{S}$. These structure tokens modify the message passing of various domains in a structurally synergistic manner, allowing the pre-trained model to extract both a holistic and domain-specific structural knowledge from a wide range of source domains.

Finally, we fuse $\Vec{H}^\mathtt{SAL}$ with $\Vec{H}^{\mathtt{FAL}}$ in Eq.~\eqref{eq.pre-train.feature-align}  to obtain the multi-domain node embedding matrix $\vec{H}$, incorporating both feature and structure alignment, as shown below.
\begin{align}\label{eq.source-fusion}
    \vec{H}^{\mathtt{AL}} =\Vec{H}^{\mathtt{FAL}}+\alpha\Vec{H}^\mathtt{SAL},
\end{align}
where $\alpha>0$ is a hyperparameter.

\stitle{Pre-training loss.}\label{sec.pre-training}
We leverage a universal task template based on subgraph similarity calculation \cite{liu2023graphprompt,yu2023generalized}, which ensures compatibility across different tasks such as node classification and graph classification. %Note that our framework is flexible with other pre-training methods. 
As demonstrated in GraphPrompt+ \cite{yu2023generalized}, prevailing contrastive pre-training objectives can be unified under this template, making them suitable choices for the pre-training loss in \model. 
%This inherent compatibility of the pre-training task could enhance the model's ability to adapt seamlessly to diverse downstream tasks. 
In general, we can adopt the following form of contrastive loss in pre-training.
\begin{align}\label{eq:generalized_loss}
     \textstyle     \bL_\text{pre}(\bO;\Theta,\bT,\Psi)=-\sum_{o\in \bO}\ln\frac{\sum_{a\in \text{Pos}_o}\exp(\mathtt{sim}(\vec{h}_{a}, \vec{h}_{o})/\tau)}{\sum_{b\in \text{Neg}_o}\exp(\mathtt{sim}(\vec{h}_{b}, \vec{h}_{o})/\tau)},
\end{align}
where $\bO$ denotes the set of observed graph element in pre-training, $a\in\text{pos}_o,b\in\text{neg}_o$ represent the positive or negative instance of $o$, respectively, and $\vec{h}_o,\vec{h}_a,\vec{h}_b$ are their corresponding embeddings. 
Furthermore, $\mathtt{sim}(\cdot,\cdot)$ is a similarity function, such as cosine similarity \cite{rahutomo2012semantic} in our implementation, and $\tau>0$ is a temperature hyperparameter. 
Note that \model\ is flexible in the materialization of $o,a,b$ to realize different contrastive losses \cite{yu2023generalized}. Our experiments adopt GraphCL \cite{you2020graph}, where \( o \) is the original graph \( G \), and \( a,b \) represent two different augmentations of  \( G \).
Hence, $\vec{h}_o,\vec{h}_a,\vec{h}_b$ are the corresponding graph embeddings, which can be obtained through a readout operation \cite{liu2023graphprompt} on the aligned node embeddings in $\mathbf{H}^{\mathtt{AL}}$.
%\( o \) represents an augmented graph \( G' \), \( a \) is the original graph \( G \), and \( b \) is a another augmented graph of \( G'' \).

The pre-training loss is optimized by updating the weights of graph encoder $\Theta$, structure tokens across all source domains %$\bT=\{\bT_{S_i}: \forall S_i \text{ s.t. } D_{S_i} \in \bD_{S}\}$, 
$\bT=\{\bT_{S_1},\ldots,\bT_{S_K}\}$, and feature alignment parameters $\Psi$.

\subsection{Cross-domain Structure Adaptation}
Beyond multi-domain pre-training, another challenge lies in cross-domain adaptation. Given a model pre-trained on graphs $\bG_S$ from source domains $\bD_S$, we aim to adapt it to a downstream task on graphs $\bG_T$ from a target domain $D_T \notin \bD_S$. 
As this work focuses on structure adaptation, we directly apply previous work  \cite{yu2024text} for feature adaptation, as outlined in Sect.~\ref{sec:problem_definition}. 

For structure adaptation, we propose \emph{dual prompts}, consisting of \emph{\op} and \emph{\cp}. On one hand, the \op\ are designed to holistically utilize the pre-trained structural knowledge from all source domains. On the other hand, the \cp\ combine multi-domain structure tokens through a learnable mixture, adapting fine-grained, domain-specific structural knowledge to the target domain.

% As both the features and structures of these domains can exhibit divergent distributions, effective integration of these multi-domain graphs requires aligning both.
% As our work focuses on structure alignment, we simply follow previous works \cite{zhao2024all,yu2024few} for feature alignment. 

% , our objective extends to bridge the gap between source and target domains. Previous works solely leverage feature adaptation, failing to narrow the structural gap \cite{zhao2024all}. Therefore, we propose a novel structure adaptation strategy and integrate feature adaptation technique.

% For feature adaptation, we first employ the same dimension alignment method as used in the pre-training phase. Given a downstream graph $G=(V,E,\Vec{X}) \in \bG_T$ from target domain $D_T$, we transform its feature matrix to $\tilde{\Vec{X}}=\mathtt{DA}_T(\Vec{X})$. We then cooperate a feature adaptation technique to align the target domain feature semantics with the source domains as follow.
% \begin{align}\label{eq.feature-emb}
%     \Vec{H}_\text{feat} = \mathtt{GE}(\mathtt{SAG}(\bG,\tilde{\Vec{X}};\Gamma);\Theta_\text{pre}),
% \end{align}
% where $\mathtt{SAG}$ represents the feature adaptation method, and $\Gamma$ denotes the learnable parameters in $\mathtt{SAG}$.

\stitle{Holistic prompts.} 
To transfer the holistic multi-domain structural knowledge to a downstream task, we propose a set of \op\ designed to align the target domain \( D_T \) with the model pre-trained on the source domains \( \bD_S \). %Given a graph \( G_T = (V_T, E_T, \Vec{X}_T) \) from \( D_T \), 
Like any pre-training framework, we encode a downstream graph $G=(V,E,\tilde{\vec{X}})$ using the pre-trained graph encoder with frozen layer-wise weights $\Theta_\text{pre}=\{\theta^1_\text{pre},\ldots,\theta^L_\text{pre}\}$. 
However, the key difference is that we inject a series of learnable vectors \( \bP_\text{hol} = \{ \Vec{p}^1_\text{hol},  \ldots, \Vec{p}^L_\text{hol} \} \) as \op\ into the downstream structure-based aggregation:
\begin{align}\label{eq.open-prompt}
    \vec{h}^l_{v} = \mathtt{Aggr}(\vec{h}^{l-1}_{v}, \{\Vec{p}^l_\text{hol} \odot \vec{h}^{l-1}_{u} : u\in\bN_v\}; \theta^l_\text{pre}),~\forall v\in V.
\end{align}
The final layer outputs a holistic node embedding matrix for the downstream graph $G$, denoted as \( \vec{H}^\text{hol}\).
%\( \vec{H}^\text{hol}=\mathtt{STACK}(\vec{\tilde{h}}_{v}:~\forall v\in G_T) \).

\stitle{Specific prompts.}
In contrast to the \op, \cp\ are designed to adapt structural knowledge specific to each source domain. Since knowledge from related source domains is likely to be more applicable, it is essential to align the target domain with different source domains to varying extents, prioritizing the most relevant ones. Consequently, we define \cp\ as \( \bP_\text{spe} = \{ \Vec{p}^1_\text{spe}, \ldots, \Vec{p}^L_\text{spe} \} \), which will also be injected into different layers of the pre-trained graph encoder. Specifically, in the $l$-th layer, \( \Vec{p}^l_\text{spe} \) is a combination of \( \{ \vec{t}^l_{S_1},\ldots,\vec{t}^l_{S_K}\} \), the pre-trained structure tokens in the corresponding layer 
%\( \bT^l_\text{pre} = \{ t^l_{S_i} \mid \forall S_i, \text{s.t.}~D_{S_i} \in \bD_{S} \} \), 
 across all source domains $D_{S_i} \in \bD_{S}$.
%which are integrated into the \( l \)-th layer of the pre-trained graph encoder. %The \cp\ serve as intermediate prompts, balancing between hard and soft prompts \cite{liu2023pre}, by utilizing pre-trained tokens alongside learnable fusion coefficients in downstream tasks. 
Formally, we define
\begin{align}\label{eq.specific-prompt-generation}\textstyle
\Vec{p}^l_\text{spe} = \sum_{i=1}^K \lambda^l_i \Vec{t}^l_{S_i},
\end{align}
where \( \Lambda^l = \{ \lambda^l_1, \ldots, \lambda^l_K \} \) are learnable coefficients. Thus, the full set of learnable parameters for the \cp\ is \( \Lambda = \{ \Lambda^1, \ldots, \Lambda^L \} \). Subsequently, \cp\ modify the structure-based aggregation in the same way as in Eq.~\eqref{eq.open-prompt}, while freezing the pre-trained weights of the graph encoder. Similarly, we denote the output node embedding matrix based on the specific prompts as \( \vec{H}^\text{spe} \).

\stitle{Prompt tuning.}
To leverage both holistic multi-domain and domain-specific structural knowledge from the pre-trained model, we fuse the output embedding matrices obtained via \op\ and \cp\ as follows.
\begin{align}\label{eq.structure-emb}
    \vec{H}^\mathtt{SAD} =\vec{H}^\text{hol}+\beta\vec{H}^\text{spe},
\end{align}
where $\beta>0$ is a hyperparameter. 
Further incorporating feature adaptation in Eq.~\eqref{eq.feature-emb}, 
we obtain the overall node embedding matrix with both feature and structure adaptations, given by
%and structure-level adapted embeddings $\vec{H}^\text{SAD}$ from Eq.~(\ref{eq.structure-emb}) as follow.
\begin{align}\label{eq.down-fusion}
    \vec{H}^\mathtt{AD} =\vec{H}^\mathtt{FAD}+\alpha\vec{H}^\mathtt{SAD}.
\end{align}
Here, $\alpha$ is the same hyperparameter used in Eq.~\eqref{eq.source-fusion}, as both share the objective of integrating the feature and structure counterparts.

For downstream node and graph classification tasks, the loss function \( \mathcal{L}_\text{down} \) is formulated based on the same task template with subgraph similarity \cite{liu2023graphprompt}, akin to the pre-training loss \( \mathcal{L}_\text{pre} \). Let \( \Omega = \{(x_1, y_1), (x_2, y_2), \ldots\} \) represent the labeled training set, where each \( x_i \) is either a node or graph instance, and \( y_i \in Y \) is its respective class from the set \( Y \). Subsequently, we optimize the following cross-domain adaptation loss:
\begin{align}\label{eq.downstream_loss}\textstyle
    \bL_\text{down}(\Omega;\bP_\text{hol},\Lambda,\Gamma)=-\sum_{(x_i,y_i)\in \Omega}\ln\frac{\exp(\text{sim}(\vec{h}_{x_i},{\vec{h}}_{y_i})/\tau )}{\sum_{y\in Y}\exp(\text{sim}(\vec{h}_{x_i},{\vec{h}}_{y})/\tau )}.
\end{align}
Here, $\vec{h}_{x_i}$ represents the adapted embedding of the node or graph $x_i$ based on $\vec{H}^\mathtt{AD}$, 
where a readout operation on $\vec{H}^\mathtt{AD}$ is required if $x_i$ is a graph. 
Additionally, ${\vec{h}}_y$ denotes the prototype embedding for class $y$, which is calculated as the average embeddings of all training instances of class $y$. %For a node $v$, $\vec{\hat{h}}_{v}$ is a row of $\Vec{\hat{H}}$. 

We outline the key steps for prompt tuning in Algorithm~\ref{alg.prompt}, Appendix~\ref{app.alg} and assess its complexity in Appendix~\ref{complexity}.

\section{Experiments}
In this section, we conduct experiments to assess the performance of \model\ and analyze its empirical results.

\begin{table}[tbp]
\center
\caption{Summary of datasets. 
\label{table.datasets}}
%\vspace{-2mm}
\resizebox{1\linewidth}{!}{%
\begin{tabular}{@{}c|rrrrrrr@{}}
\toprule
	& \makecell[c]{Nodes} & \makecell[c]{Edges} & \makecell[c]{Feature\\dimension} & \makecell[c]{Node\\classes} & \makecell[c]{Avg.\\nd} &\makecell[c]{Avg.\\spl} &\makecell[c]{Avg.\\cc}\\
\midrule
     Cora & 2,708 & 10,556 & 1,433 & 7 & 3.89 & 6.30
 & 0.24 \\
     Citeseer & 3,327 & 9,104 & 3,703 & 6 & 2.73
&9.31 & 0.14 \\ 
     Pubmed & 19,717 & 88,648 & 500 & 3 & 4.49
&6.33   & 0.06\\
     Photo & 7,650 & 238,162 & 745 & 8 & 31.13
&4.05   & 0.40 \\
     Computers & 13,752 & 491,722 & 767 & 10 & 35.75
&3.38   & 0.34\\
     Facebook & 22,470 & 342,004 & 128 & 4 & 15.22
&4.97   &  0.35 \\
     LastFM & 7,624 & 55,612 & 128 & 18 & 7.29
& 5.23 & 0.21

\\
 \bottomrule
\end{tabular}}
   \parbox{1\linewidth}{\footnotesize nd: node degree, spl: shortest path length \cite{borgwardt2005shortest}, cc: clustering coefficient \cite{giatsidis2014corecluster}.}
\end{table}

\begin{table*}[tbp] % [!t]
    \centering
    \small
     \addtolength{\tabcolsep}{0.7mm}
    \caption{Accuracy (\%) of one-shot \emph{node classification} with standard deviations. Each column represents a target domain, using other columns as source domains.  The best method in each column is bolded, and the runner-up is underlined.
    }
   % \vspace{-2mm}
    \label{table.node-classification}%
    %\resizebox{1\linewidth}{!}{%
    \begin{tabular}{c|c|c|c|c|c|c|c}
    \toprule
   {{Method }\textbackslash{ Target domain}}   & Cora & Citeseer & Pubmed & Photo & Computers & Facebook & LastFM
      \\\midrule\midrule
    \method{GCN} 
    & 29.53 $\pm$ \phantom{0}7.56 
    & 26.29 $\pm$ \phantom{0}6.50  
    & 23.32 $\pm$ 11.56  
    & 26.96 $\pm$ 12.94 
    & 24.40 $\pm$ \phantom{0}5.62 
    & 20.45 $\pm$ \phantom{0}5.62 
    & \phantom{0}9.21 $\pm$ \phantom{0}3.11   
\\ 
    \method{GAT} 
    & 24.27 $\pm$ \phantom{0}9.26  
    & 21.56 $\pm$ \phantom{0}8.09    
    & 22.28 $\pm$ \phantom{0}9.76   
    & 17.85 $\pm$ 10.22 
    & 23.03 $\pm$ 12.12 
    & 29.27 $\pm$ \phantom{0}6.47   
    & \phantom{0}9.01 $\pm$ \phantom{0}2.61
 
\\\midrule
    \method{DGI}
    & 33.40 $\pm$ 10.48  
    & 25.80 $\pm$ \phantom{0}8.27
    & 47.22 $\pm$ \phantom{0}9.50  
    & 30.89 $\pm$ 10.54  
    & 25.75 $\pm$ 12.45  
    & 34.36 $\pm$ \phantom{0}9.57 
    & 14.14 $\pm$ \phantom{0}6.31
\\
    \method{GraphCL}
    & 27.72 $\pm$ \phantom{0}9.37   
    & 35.02 $\pm$ \phantom{0}8.46  
    & \underline{48.89} $\pm$ \phantom{0}9.03  
    & 34.78 $\pm$ 11.56  
    & {23.79} $\pm$ 12.28 
    & 34.85 $\pm$  \phantom{0}7.07 
    & 18.93 $\pm$  \phantom{0}7.32 
    
\\%\midrule    
    \method{GPPT}
    & 27.18 $\pm$ \phantom{0}4.88	
    & 25.90 $\pm$ \phantom{0}4.68 
    & 39.82 $\pm$ \phantom{0}8.79 
    & 31.58 $\pm$ 10.27  
    & 19.94 $\pm$ \phantom{0}9.61
    & 34.73 $\pm$ \phantom{0}3.99 
    & 20.98 $\pm$ \phantom{0}3.98
\\
    \method{GraphPrompt}
    & 28.26 $\pm$ 12.68
    & 32.51 $\pm$ \phantom{0}8.73
    & 47.47 $\pm$ \phantom{0}9.15
    & 48.11 $\pm$ \phantom{0}9.89  
    & 42.82 $\pm$ 11.67 
    & 40.44 $\pm$ \phantom{0}9.68
    & 19.84 $\pm$ \phantom{0}7.23 
\\
    \method{GPF}
    & 32.17 $\pm$ \phantom{0}6.56
    & \underline{36.79} $\pm$ \phantom{0}7.70 
    & 41.28 $\pm$ \phantom{0}8.14 
    & 47.47 $\pm$ \phantom{0}8.19  
    & 35.75 $\pm$ \phantom{0}7.12 
    & 40.45 $\pm$ \phantom{0}6.34
    & 27.26 $\pm$ \phantom{0}5.50
\\\midrule
    \method{Hassani}
    & 33.35 $\pm$ \phantom{0}6.93
    & 33.66 $\pm$ \phantom{0}7.24
    & 39.87 $\pm$ \phantom{0}8.16
    & 48.48 $\pm$ \phantom{0}7.07
    & 39.99 $\pm$ \phantom{0}7.91
    & 37.70 $\pm$ \phantom{0}5.79
    & 27.16 $\pm$ \phantom{0}4.94
    
\\\midrule

    \method{GCOPE}
    & \underline{35.62} $\pm$ 11.93	
    & \textbf{38.33} $\pm$ \phantom{0}9.28  
    & 45.38 $\pm$ \phantom{0}9.87  
    & \underline{52.87} $\pm$ \phantom{0}9.19
    & \underline{45.65} $\pm$ 10.69 
    & \underline{40.63} $\pm$ \phantom{0}8.50
    & \underline{28.84} $\pm$ \phantom{0}7.59
 \\
  
      \method{\model}
    & \textbf{47.80} $\pm$ 11.88  
    & 36.38 $\pm$ \phantom{0}9.10
    & \textbf{50.25} $\pm$ 10.43 
    & \textbf{58.71} $\pm$ \phantom{0}8.69 
    & \textbf{48.22} $\pm$ \phantom{0}8.17 
    & \textbf{42.70} $\pm$ \phantom{0}8.73
    & \textbf{33.36} $\pm$ \phantom{0}8.11  
    
\\    \bottomrule
        \end{tabular}%}
        \\
%   \parbox{1\linewidth}{\centering In each column, the best-performing method is highlighted in bold, and the runner-up is underlined. Table~\ref{table.graph-classification} follows the same format.}
\end{table*}

\begin{table*}[tbp] % [!t]
    \centering
    \small
     \addtolength{\tabcolsep}{0.7mm}
    \caption{Accuracy (\%) of one-shot \emph{graph classification} with standard deviations. Each column represents a target domain, using other columns as source domains.  The best method in each column is bolded, and the runner-up is underlined.
    }
    % \vspace{-2mm}
    \label{table.graph-classification}%
    %\resizebox{1\linewidth}{!}{%
    \begin{tabular}{c|c|c|c|c|c|c|c}
    \toprule
   {Method }\textbackslash{ Target domain}   & Cora & Citeseer & Pubmed & Photo & Computers & Facebook & LastFM
      \\\midrule\midrule
     \method{GCN} 
    & 30.64 $\pm$ 10.31 
    & 26.90 $\pm$ \phantom{0}7.15 
    & 38.84 $\pm$ 11.82 
    & 15.60 $\pm$ \phantom{0}8.77   
    & 21.94 $\pm$ 14.51 
    & 31.33 $\pm$ \phantom{0}9.47  
    & 28.83 $\pm$ \phantom{0}9.60
    
\\ 
    \method{GAT} 
    & 27.80 $\pm$ \phantom{0}7.85  
    & 27.50 $\pm$ \phantom{0}7.13    
    & 21.66 $\pm$ \phantom{0}8.70    
    & 15.74 $\pm$ \phantom{0}7.62 
    & 16.02 $\pm$ 13.46 
    & 21.20 $\pm$ \phantom{0}7.31  
    & 27.80 $\pm$ \phantom{0}7.85
 
\\\midrule
    \method{InfoGraph}
    & 34.98 $\pm$ 10.15 
    & 35.87 $\pm$ \phantom{0}9.84
    & 48.67 $\pm$ 12.29  
    & 25.70 $\pm$ 11.73  
    & 19.02 $\pm$ 14.09  
    & 31.26 $\pm$ \phantom{0}9.65 
    & 23.29 $\pm$ \phantom{0}7.99
\\
    \method{GraphCL}
    & \underline{42.70} $\pm$ 10.64
    & 36.66 $\pm$ \phantom{0}8.67
    & 47.53 $\pm$ 11.52  
    & 33.07 $\pm$ 12.31  
    & 16.02 $\pm$ 13.47 
    & 21.99 $\pm$ 13.00
    & 21.30 $\pm$ 10.45
\\%\midrule    
    \method{GraphPrompt}
    & 37.38 $\pm$ 14.03	
    & 36.66 $\pm$ \phantom{0}9.19  
    & \textbf{49.55} $\pm$ 10.25 
    & 50.79 $\pm$ 12.31
    & 43.09 $\pm$ 11.45 
    & \underline{41.71} $\pm$ 10.61
    & 32.62 $\pm$ \phantom{0}8.54
\\
    \method{GPF}
    & 39.62 $\pm$ \phantom{0}8.52	
    & 36.73 $\pm$ \phantom{0}7.66 
    & 45.08 $\pm$ 10.36 
    & 47.57 $\pm$ 10.16 
    & 35.70 $\pm$ \phantom{0}8.71  
    & 34.84 $\pm$ \phantom{0}5.14  
    & 34.31 $\pm$ \phantom{0}7.05

\\\midrule
    \method{Hassani}
    & 36.86 $\pm$ 10.74
    & 35.78 $\pm$ \phantom{0}8.80
    & 43.97 $\pm$ 13.27
    & 41.55 $\pm$ 13.08
    & 29.49 $\pm$ 13.86
    & 35.57 $\pm$ \phantom{0}9.00
    & 25.39 $\pm$ \phantom{0}8.14
\\\midrule

    \method{GCOPE}
    & 38.85 $\pm$ 10.99	
    & \textbf{39.93} $\pm$ \phantom{0}9.82 
    & 47.05 $\pm$ 11.74
    & \underline{53.93} $\pm$ \phantom{0}9.74  
    & \underline{45.60} $\pm$ 10.96 
    & 40.26 $\pm$ \phantom{0}9.53
    & \underline{34.68} $\pm$ \phantom{0}7.70
 \\
  
      \method{\model}
    & \textbf{55.35} $\pm$ 13.62  
    & \underline{38.75} $\pm$ \phantom{0}9.40
    & \underline{48.69} $\pm$ 10.16 
    & \textbf{58.75} $\pm$ 11.67 
    & \textbf{48.72} $\pm$ 11.18 
    & \textbf{43.71} $\pm$ \phantom{0}9.54
    & \textbf{48.28} $\pm$ \phantom{0}9.72 
\\    \bottomrule
        \end{tabular}%}
\end{table*}

\subsection{Experimental Setup}
\stitle{Datasets.}
We conduct experiments on seven benchmark datasets.
(1) \emph{Cora} \cite{mccallum2000automating}, (2) \emph{Citeseer} \cite{sen2008collective} and (3) \emph{Pubmed} \cite{sen2008collective} are scientific paper citation networks from different fields, including computer science and biomedical research. Nodes represent academic publications and edges denote citation relationships.
(4) \emph{Photo} \cite{shchur2018pitfalls} and (5) \emph{Computers} \cite{mcauley2015image} are both e-commerce networks from Amazon in different categories, namely, photography and computer related products. Nodes represent products and edges signify frequent co-purchases between products.
(6) \textit{Facebook} \cite{rozemberczki2021multi} is a Web graph, where nodes represent official Facebook pages while the links are mutual likes between these pages.
(7) \textit{LastFM} \cite{rozemberczki2020characteristic} is a social network, where nodes denote users and edges represent interactions such as follower relationships. Note that each domain comprises a single graph. We summary these datasets in Table~\ref{table.datasets} and present additional details in Appendix~\ref{app.dataset}.

\stitle{Setup of pre-training and downstream tasks.}
Following previous work \cite{zhao2024all,yu2024text}, we treat each dataset as a distinct domain. 
Among the seven datasets (or domains), we use each of them as the target domain while leveraging the remaining six as  source domains for pre-training.

On each target domain, we conduct $m$-shot \textit{node classification} and \textit{graph classification}, where $m$ labeled nodes or graphs per class are randomly selected for downstream prompt tuning. Given that each dataset comprises a single graph, performing graph classification on whole graphs is not feasible. Therefore, following previous works \cite{lu2021learning,yu2023hgprompt,yu2024non}, we generate a series of graphs by constructing ego-networks centered on the labeled nodes within each dataset, and set up graph classification on these ego-networks, with each network labeled according to its central node.
 Note that the graph encoder is pre-trained only once for each set of source domains, and subsequently utilized across all downstream tasks.
We generate 100 $m$-shot tasks for both node classification and graph classification by repeatedly sampling $m$ labeled nodes or graphs per class for 100 times. Each task is executed with five different random seeds, leading to a total of 500 outcomes for each classification type. 
We use accuracy as the evaluation metric, as each task is class-balanced \cite{wang2020graph,liu2021relative,liu2023graphprompt,yu2023generalized}, and report the average accuracy and standard deviation over these 500 outcomes.

\stitle{Baselines.}
We compare the performance of \model\ against state-of-the-art methods in four broad groups, as follows.

(1) \emph{End-to-end graph neural networks}: GCN \cite{kipf2016semi} and GAT \cite{velivckovic2017graph} aggregate information from neighboring nodes to update node representations. For each task, they are trained from scratch in a supervised fashion without pre-training.

(2) \emph{Graph pre-training models}: DGI \cite{velivckovic2018deep}, InfoGraph \cite{sun2019infograph}\footnote{Original DGI only operates at the node level, while InfoGraph extends it to the graph level. We apply DGI to node classification, and InfoGraph to graph classification.} and GraphCL \cite{you2020graph} first pre-train a graph encoder to capture the inherent properties of the graphs, and then fine-tune a classifier on the downstream task while freezing the pre-trained model.  GPPT\footnote{GPPT is tailored for node classification task and is not applicable to graph classification. Thus, in our experiments, we only use GPPT for node classification.} \cite{sun2022gppt}, GPF \cite{fang2022universal} and GraphPrompt \cite{liu2023graphprompt}  employ a universal task template to unify self-supervised pre-training and downstream tasks, and tune a single prompt on downstream tasks. 

(3) \textit{Graph cross-domain models}: Hassani \cite{hassani2022cross} pre-trains a GNN on a single source domain by incorporating both contextual and topological views, which facilitates cross-domain adaptation for downstream tasks.

(4) \emph{Multi-domain pre-training models}: GCOPE \cite{zhao2024all} performs multi-domain pre-training, and subsequently adapts to cross-domain tasks through either fine-tuning a classification head or prompt tuning. We opt for fine-tuning as it yields superior performance. 

Note that the above graph pre-training and cross-domain approaches are originally designed for pre-training on a single source domain. For a fair comparison, we directly merge the multi-domain graphs and apply dimension alignment for them, as in \model.
Further descriptions of the baselines are provided in Appendix~\ref{app.baselines}, with implementation details in Appendix~\ref{app.parameters}.

\subsection{Performance Evaluation}\label{sec.exp.per}
We first compare \model\ and the baseline methods on one-shot node and graph classification tasks, and then investigate the effect of increasing the number of shots.

\stitle{One-shot performance.}\label{exp.main}
Tables~\ref{table.node-classification} and~\ref{table.graph-classification} show the results of one-shot node and graph classification tasks. 
We observe that, first, \model\ achieves outstanding performance in both node and graph classification across various target domains, demonstrating the effectiveness of our proposed structure tokens in multi-domain pre-training and dual prompts in cross-domain adaptation. We defer analysis of the quantitative contributions of these components to the ablation studies in Sect.~\ref{sec.ablation}.
Second, another text-free multi-domain pre-training method, GCOPE, significantly lags behind \model\ because it only performs alignment and adaptation on feature and  homophily patterns, without accounting for structural differences across domains.
This further emphasizes the importance of our structure tokens and dual prompts.
Third, graph pre-training methods generally outperform the end-to-end GCN and GAT, showcasing the benefits of pre-training on unlabeled graphs.

\begin{figure}[tbp]
\centering 
\includegraphics[width=1\linewidth]{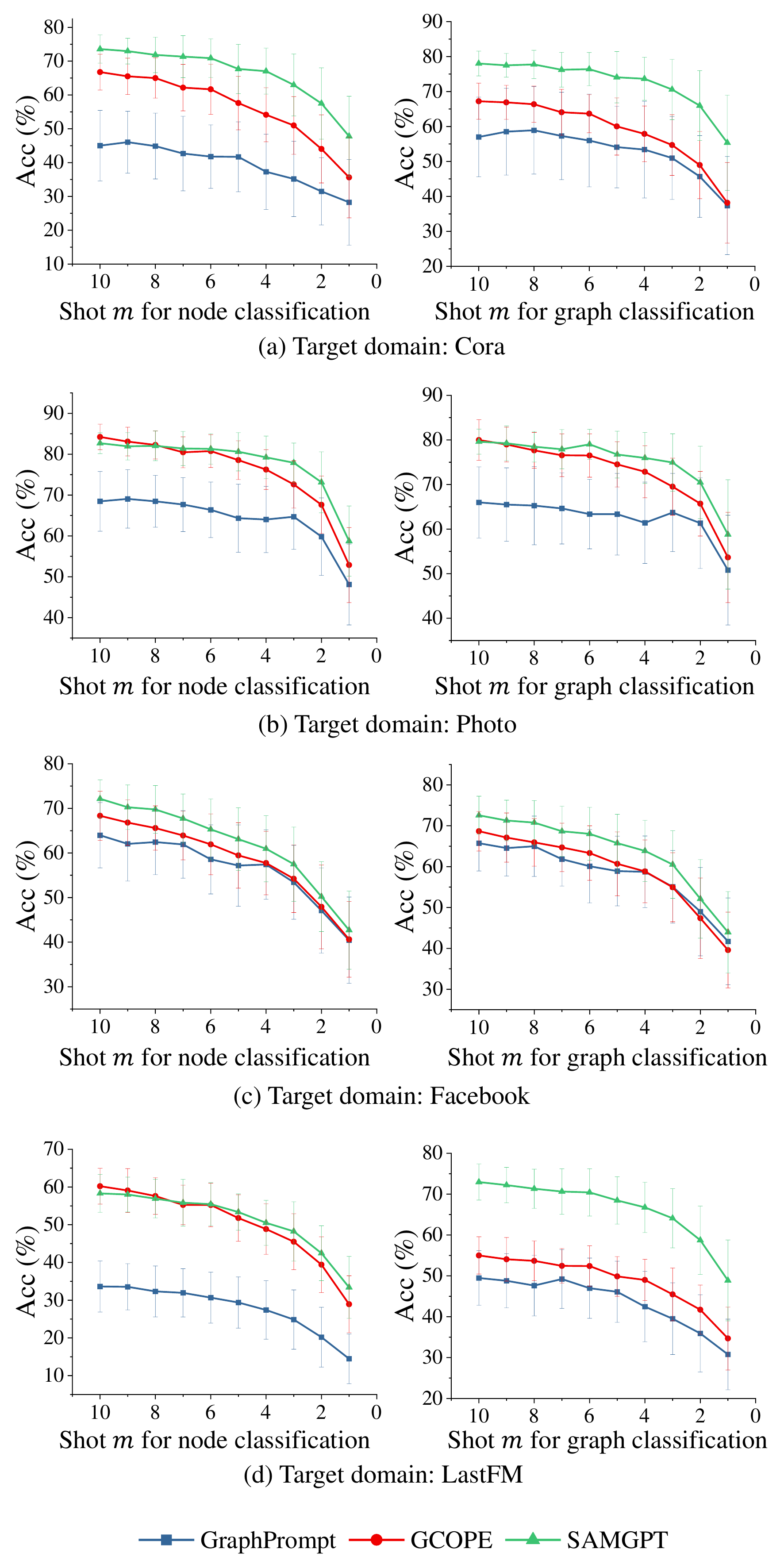}%
 \vspace{-3mm}%
\caption{Impact of number of shots on node and graph classification on four target domains.}
\label{fig.fewshot}
\end{figure}

\stitle{Few-shot performance.}
To evaluate the performance of \model\ with more labeled data, we vary the number of shots, $m$, in both node and graph classification tasks. We compare \model\ to two competitive baselines, \method{GraphPrompt} and \method{GCOPE}, with results reported in Fig.~\ref{fig.fewshot}, where error bars represent the standard deviation. We observe that \model\ consistently outperforms the baselines in low-shot settings (\eg, $m\le 5$). When further increasing the number of shots, \model\ still performs best in general, although it may be on par with \method{GCOPE} in some cases when $m$ approaches 10. This is not surprising, since the advantage of \model\ may diminish as more supervision becomes available.

%Its performance is  demonstrating the robustness of \model\ when 

% \begin{figure}[t]
% \centering 
% \includegraphics[width=0.5\linewidth]{figures/data_ablation.pdf}
% \caption{Data ablation study with an increasing number of source domains.}
% \label{fig.data-ablation}
% \end{figure}

\subsection{Ablation Studies}\label{sec.ablation}
To understand the impact of each component in \model, we perform two ablation studies.

\stitle{Data ablation.}
We evaluate the impact of incorporating more source domains by incrementally adding \textit{Citeseer}, \textit{LastFM}, \textit{Photo}, and \textit{Facebook}, in this order, to the pre-training, while fixing \textit{Cora} as the target domain. 
We present one-shot node classification performance of \model\ and two competitive baselines, namely, \method{GraphPrompt} and \method{GCOPE}, in Table~\ref{table.data-ablation}. Across the columns, 1 represents using \textit{Citeseer} as the single source domain, while 2 represents using \textit{Citeseer} and \textit{LastFM} as the source domains, \etc 

\begin{table}[tbp] % [!t]
    \centering
    \small
     \addtolength{\tabcolsep}{-0.5mm}
    \caption{Data ablation study with an increasing number of source domains, while fixing \emph{Cora} as the target domain.
    }
   %  \vspace{-2mm}
    \label{table.data-ablation}%
    \resizebox{1\linewidth}{!}{%
    \begin{tabular}{l|cccc}
    \toprule
   \multirow{2}*{Method} &\multicolumn{4}{c}{Number of source domains}\\   & 1 & 2 & 3 & 4
      \\\midrule%\midrule
    \method{GraphPrompt}
    & 35.53\text{\scriptsize ±12.06}
    & 37.13\text{\scriptsize ±11.79}
    & 36.90\text{\scriptsize ±11.23}
    & 38.54\text{\scriptsize ±11.84}  
\\
    \method{GCOPE}
    & 39.47\text{\scriptsize ±12.14}
    & 36.63\text{\scriptsize ±\phantom{0}9.46}
    & 35.28\text{\scriptsize ±11.99}
    & 38.61\text{\scriptsize ±12.74}
 \\
      \method{\model}
    & \textbf{40.43}\text{\scriptsize ±11.00}
    & \textbf{41.97}\text{\scriptsize ±11.01}
    & \textbf{42.30}\text{\scriptsize ±11.56}
    & \textbf{45.95}\text{\scriptsize ±12.96}
\\    \bottomrule
        \end{tabular}}
\end{table}

\begin{table*}[tbp]
    \centering
    \small
    \addtolength{\tabcolsep}{0.5mm}
    \caption{Model ablation study on key components of \model.}
    % \vspace{-2mm}
    \label{table.ablation}%
    %\resizebox{1\linewidth}{!}{%
    \begin{tabular}{c|ccc|ccc|ccc}
    \toprule
    \multirow{2}*{Methods}
    & Structure & Holistic & Specific &\multicolumn{3}{c|}{Target domain for node classification} &\multicolumn{3}{c}{Target domain for graph classification}\\
    &tokens &prompts &prompts & Cora & Photo & Facebook & Cora & Photo & Facebook\\
    \midrule%\midrule
    \method{Variant 1}
    & $\times$ & $\times$ & $\times$ 
    &36.36 $\pm$ 12.71 & 49.10 $\pm$ \phantom{0}9.94 & 35.36 $\pm$ \phantom{0}9.06
    & 45.44 $\pm$ 13.47 & 52.45 $\pm$ 12.37 & 38.74 $\pm$ 10.26
    \\
    \method{Variant 2}
    & $\times$ & $\times$ & $\checkmark$ 
    &40.62 $\pm$ 11.79  & 56.23 $\pm$ \phantom{0}9.04  & 39.80 $\pm$ 10.39 
    &45.63 $\pm$ 13.52  & 57.78 $\pm$ 11.64  & 42.22 $\pm$ 10.95    \\ 
    \method{Variant 3} 
    & $\checkmark$ & $\times$ & $\times$
    & 44.26 $\pm$ 10.92  & 56.61 $\pm$ 10.14  & 41.11 $\pm$ \phantom{0}8.34 
    & 52.88 $\pm$ 12.25  & 58.14 $\pm$ 12.01  & 43.12 $\pm$ \phantom{0}9.76    \\  
    \method{Variant 4}
    & $\checkmark$ & $\checkmark$ & $\times$ 
    &46.10 $\pm$ 12.02  & 57.76 $\pm$ 10.00  & 40.46 $\pm$ \phantom{0}8.89
    &54.52 $\pm$ 14.32  & 58.12 $\pm$ 12.30  & 43.15 $\pm$ 10.12    \\ 
    \method{\model}
    & $\checkmark$ & $\checkmark$ & $\checkmark$
    &\textbf{47.80} $\pm$ 11.88 &
    \textbf{58.71} $\pm$ \phantom{0}8.69 &
    \textbf{42.70} $\pm$ \phantom{0}8.73 &
    \textbf{55.35} $\pm$ 13.62 & \textbf{58.75} $\pm$ 11.67&
    \textbf{43.71} $\pm$ \phantom{0}9.54 \\
    \bottomrule
    \end{tabular}%}
\end{table*}

We make the following observations. First, \model\ is superior across different numbers of source domains, demonstrating its robustness to varying  configurations of the source domains. Second, both \method{GraphPrompt} and \method{GCOPE} often perform worse as more datasets are added due to the negligence of  structural discrepancies in various domains. In contrast, \model\ exhibits consistent improvement with the addition of more source domains, validating the effectiveness of our structure alignment and adaptation.

\stitle{Model ablation.}
We analyze variants of \model\ by removing structure tokens, \op\ and \cp. We report the results of these variants and \model\ in Table~\ref{table.ablation}. 
Note that Variant 1, which lacks our structural alignment design, is equivalent to the feature alignment method MDGPT \cite{yu2024text}.

The results confirm that each component plays an important role.
First, the use of structure tokens is essential. Notably, Variant 3 consistently outperforms Variant 1 and 2, both of which do not employ structure tokens, demonstrating the effectiveness of structure tokens in aligning multi-domain structural knowledge.
Second, removing \cp\ leads to a drop in performance, evident from the superior accuracy of Variants 2 over Variant 1, and \model\ over Variant 4. This indicates the significance of leveraging source domain-specific structural knowledge for effective cross-domain adaptation. 
Third, \op\ prove to be useful, as Variant 4 often outperforms Variant 3, highlighting the significance of incorporating holistic multi-domain structural information via \op.
These key components together enable \model\ to achieve optimal performance.

\begin{table}[tb]
    \centering
    \small
    \addtolength{\tabcolsep}{-.1mm}
    \caption{Analysis of one-shot node classification on homophilic and heterophilic graphs.}
   %  \vspace{-2mm}
    \label{table.homo-hetero}%
    %\resizebox{1\linewidth}{!}{%
    \begin{tabular}{@{}c|c|ccc@{}}
    \toprule
    Target & \multirow{2}*{Source domains} & \multicolumn{3}{c}{Accuracy (\%)} \\
    domain &  & \method{GraphPrompt} & GCOPE & \model  \\
    \midrule%\midrule
    Squi.
    & Cham., Corn., Cora &  
    18.98\text{\scriptsize ±4.89} & 18.98\text{\scriptsize ±4.75} & \textbf{20.43}\text{\scriptsize ±4.75} \\
    Corn.
    & Squi., Cham., Cora &
    29.67\text{\scriptsize ±8.36}  & 27.19\text{\scriptsize ±8.51} & \textbf{32.57}\text{\scriptsize ±8.68}\\
    Cham.
    & Squi., Corn., Cora &  
    23.28\text{\scriptsize ±4.63}  & 23.24\text{\scriptsize ±4.50} & \textbf{23.89}\text{\scriptsize ±4.91}\\   
    Facebook
    & Squi., Cora, Photo &  
    32.22\text{\scriptsize ±6.91}  & 35.81\text{\scriptsize ±7.89} & \textbf{41.10}\text{\scriptsize ±9.38}\\
    \bottomrule
    \end{tabular}\\%}
       \parbox{1\linewidth}{\footnotesize \ Squi., Cham., Corn. are short for Squirrel, Chameleon, and Cornell, respectively.}
\end{table}

\subsection{Homophily Sensitivity}\label{sec.hetero}
Apart from feature and structural differences, graphs also exhibit varying homophily and heterophily patterns based on whether linked nodes share the same attribute \cite{ma2021homophily,zhu2020beyond,yu2024non}. To further assess the robustness of \model\ across domains with varying homophily patterns, we conduct one-shot node classification on  homophilic (\textit{Cora}, \textit{Photo}, \textit{Facebook}) and heterophilic (\textit{Chameleon}, \textit{Cornell} and \textit{Squirrel}) graphs. Details about the heterophilic datasets are presented in Appendix~\ref{app.hetero}. 

We report the results in Table~\ref{table.homo-hetero} and observe that \model\ consistently surpasses \method{GraphPrompt} and \method{GCOPE}, regardless of whether the source or target domains are homophilic or heterophilic. 
These results further validate the efficacy of \model, demonstrating its ability to leverage multi-domain knowledge across a wide variety of graph domains. Note that we focus on the node classification task here, as homophily is defined based on node attributes, which directly impacts node-level tasks.

% \begin{figure*}[t]
% \centering
% \begin{minipage}[b]{0.36\textwidth}
% \centering
% \includegraphics[width=0.8\linewidth]{figures/data_ablation.pdf}
% \vspace{-2mm}
% \caption{Data ablation study with an increasing number of source domains.}\vspace{-2mm}
% \label{fig.ablation}
% \end{minipage}%
% \hspace{5.4mm}%
% \begin{minipage}[b]{0.6\textwidth}
% \centering
% \includegraphics[width=0.95\linewidth]{figures/alpha.pdf}
% \vspace{-5mm}
% \caption{Sensitivity study of $\alpha$.}\vspace{-2mm}
% \label{fig.para}
% \end{minipage}
% \end{figure*}

\section{Conclusions}
In this paper, we proposed \model, a graph foundation model with structure alignment for text-free graphs, supporting both multi-domain graph pre-training and cross-domain adaptation. In the pre-training phase, \model\ utilizes a series of structure tokens to harmonize the structural distributions across multiple source domains and to extract multi-domain structural knowledge. For downstream cross-domain adaptation, \model\ employs dual prompts to tailor pre-trained holistic and domain-specific structural knowledge to the target domain. We conducted extensive experiments on seven benchmark datasets, demonstrating that \model\ significantly outperforms various state-of-the-art baseline methods.

% \begin{acks}
% \end{acks}

%\newpage

\section*{Acknowledgments}
This research / project is supported by the Ministry of Education, Singapore under its Academic Research Fund (AcRF) Tier 1 grant (22-SIS-SMU-054) and Tier 2 grant (Proposal ID: T2EP20122-0041). Any opinions, findings and conclusions or recommendations expressed in this material are those of the author(s) and do not reflect the views of the Ministry of Education, Singapore. 

\clearpage
%\newpage
\bibliographystyle{ACM-Reference-Format}
\bibliography{references}
\balance

% \newpage
\appendix
\section*{Appendices}
\renewcommand\thesubsection{\Alph{subsection}}
\renewcommand\thesubsubsection{\thesubsection.\arabic{subsection}}

\subsection{Algorithm}\label{app.alg}
Our algorithm consists of two stages, multi-domain graph pre-training and downstream adaptation.

In the multi-domain pre-training phase,  we first apply the dimension alignment function, $\mathtt{DAL}$, to align feature dimensions from different source domains by Eq.~\eqref{eq.pre-train.dim-align}. Then, we use a feature alignment method to unify feature semantic spaces by Eq.~\eqref{eq.pre-train.feature-align}. For structure alignment, we inject source domain-specific structure tokens into each layer of the graph encoder by Eq.~\eqref{eq:structure_aggregation}. Finally, we fuse the feature aligned embedding and structure aligned embedding by Eq.~\eqref{eq.source-fusion}, and optimize the pre-training loss by Eq.~\eqref{eq:generalized_loss}.

We further present the key steps for cross-domain adaptation in Algorithm~\ref{alg.prompt}. In lines 3--4, we align target domain feature dimensions with source domains. In lines 6--7, we integrate the feature adaptation method to generate feature-level adapted embeddings. In lines 8--21, we employ dual prompts to adapt structural prior knowledge to the target domain. Specifically, we first inject \op\ to modify the structure-based aggregation in each layer of the graph encoder for holistic knowledge adaptation (lines 9--12). Then, we generate \cp\ by fusing the pre-trained structure tokens (lines 13--15), and utilize \cp\ for domain-specific knowledge adaptation (lines 17--19). We obtain structure-level adapted embeddings by fusing holistic and domain-specific embeddings (lines 20--21), and generate final embeddings by aggregating feature- and structure-level adapted embeddings (lines 22--23). Finally, we update the embeddings for the prototypical instances based on the labeled samples in the task and optimize \op, $\Lambda$ and $\Gamma$ (lines 24-28). %Note that updating prototypical is required exclusively for classification tasks.

\begin{algorithm}[b]
\small
\caption{\textsc{Cross-domain Adaptation for \model}}
\label{alg.prompt}
\begin{algorithmic}[1]
    \Require Pre-trained graph encoder $\mathtt{GE}$ with parameters $\Theta_\text{pre}$, pre-trained structure tokens $\bT_\text{pre}$, target domain dimension alignment function $\mathtt{DA}_T(\cdot)$, and feature adaptation function $\mathtt{FAD}(\cdot)$
    \Ensure Optimized holistic prompts $\bP_\text{hol}$, coefficients $\Lambda$, and feature adaptation parameters $\Gamma$
    \While{not converged} 
        \For{each graph $G_T=(V_T, E_T, \vec{X}_T)$ in target domain $D_T$}
            \State \slash* Target domain feature dimensions alignment by Eq.~(\ref{eq.target-dimension}) *\slash
            \State $\tilde{\Vec{X}}\leftarrow \mathtt{DAL}_T(\Vec{X})$
            \State $\bP_\text{hol}$, $\Lambda, \Gamma \leftarrow$ initialization
                \State \slash* Feature adaptation by Eq.~(\ref{eq.feature-emb}) *\slash
                \State $\Vec{H}^\mathtt{FAD} \leftarrow \mathtt{GE}(\mathtt{FAD}(\bG,\tilde{\Vec{X}};\Gamma);\Theta_\text{pre})$
                \State \slash* Structure alignment by dual prompts *\slash
                %\State \slash* Adaptation of holistic structural prior knowledge *\slash
                \State \slash* Modification to $\mathtt{GE}$ via \op\ by Eq.~(\ref{eq.open-prompt}) *\slash
                \For{each layer in $\mathtt{GE}$}
                    \State $\vec{h}^l_{v} \leftarrow \mathtt{Aggr}(\vec{h}^{l-1}_{v}, \{\Vec{p}^l_\text{hol} \odot \vec{h}^{l-1}_{u} : u\in\bN_v\}; \theta^l_\text{pre}),~\forall v\in G_T$
                \EndFor
                \State $\vec{H}^\text{hol}\leftarrow \mathtt{STACK}(\{\vec{h}_{v}:~\forall v\in G_T\})$
                \State \slash* Generation of \cp\ by Eq.~(\ref{eq.specific-prompt-generation}) *\slash
                \For{$\Vec{p}^l_\text{spe}~\text{in}~\bP_\text{spe}$}
                    \State $\Vec{p}^l_\text{spe} \leftarrow \sum_{i=1}^K \lambda^l_i \Vec{t}^l_{S_i}$
                \EndFor
                \State \slash* Modification to $\mathtt{GE}$ via \cp *\slash
                \For{Each layer in $\mathtt{GE}$}            
                    \State $\vec{\tilde{h}}^l_{v} \leftarrow \mathtt{Aggr}(\vec{\tilde{h}}^{l-1}_{v}, \{\Vec{p}^l_\text{spe} \odot \vec{\tilde{h}}^{l-1}_{u} : u\in\bN_v\}; \theta^l_\text{pre}),~\forall v\in G_T$
                \EndFor
                \State $\vec{H}^\text{spe}\leftarrow \mathtt{STACK}(\{\vec{\tilde{h}}_{v}:~\forall v\in G_T\})$
                \State \slash* Fusion of dual-prompt tuned embeddings by Eq.~(\ref{eq.structure-emb}) *\slash
                \State $\vec{H}^\text{SAD} \leftarrow \vec{H}^\text{hol}+\beta\vec{H}^\text{spe}$
                \State \slash* Fusion of adapted embeddings by Eq.~(\ref{eq.down-fusion}) *\slash
                \State $\vec{H}^\mathtt{AD} \leftarrow \vec{H}^\text{FAD}+\alpha\vec{H}^\text{SAD}$
                \State \slash* Update prototypical nodes *\slash
                    \For{each class $y$} 
                        \State ${\vec{h}}_{y} \leftarrow \textsc{Mean}(\{\vec{h}_{x}$: instance $x$ belongs to class $y$\})
                    \EndFor
                \State \slash* Optimizing $\bP_\text{hol}$, $\Lambda$, and $\Gamma$ *\slash
                \State Calculate $\bL_\text{down}(\Omega;\bP_\text{hol},\Lambda,\Gamma)$ by Eq.~(\ref{eq.downstream_loss})
        \EndFor
    \EndWhile 
    \State \Return $\bP_\text{hol}$, $\Lambda$, and $\Gamma$
\end{algorithmic}
\end{algorithm}

\subsection{Complexity Analysis}\label{complexity}
For a downstream graph \( G_T = (V_T, E_T, \vec{X}_T) \) from the target domain \( D_T \), the computational process of structure adaptation involves injecting \op\ and \cp\ into the pre-trained GNN to modify the encoding of nodes. 

In a standard GNN, each node aggregates messages from its neighbors in each layer. Assuming that the aggregation involves at most $n$ neighbors, the complexity of calculating node embeddings over $L$ layers is $O(n^L \cdot |V_T|)$. One one hand, \op\ are directly injected into each layer of the GNN, increasing the complexity by $O(L \cdot |V_T|)$. On the other hand, \cp\ are first generated by the pre-trained structure tokens from $K$ source domains, with an additional complexity of $O(L \cdot K)$. Then, \cp\ modify the structure-base aggregation with a complexity of $O(L \cdot |V_T|)$. Since \op\ and \cp\ modifies the node encoding phase separately, the overall complexity would increase by $(L \cdot |V_T| + L \cdot K)$.

Thus, the encoding time by the pre-trained GNN still dominates the overall complexity, as $O(n^L \cdot |V_T|)$ typically exceeds $O(L \cdot |V_T|+ L\cdot K)$ given that $n^L > L$ and $|V_T| > K$ in general. In other words, our structural adaptation adds a marginal overhead to the pre-trained graph encoder.

\subsection{Further Descriptions of Datasets} \label{app.dataset}
In this section, we provide more comprehensive descriptions of the benchmark datasets used in our experiments, as summarized in Table~\ref{table.datasets},
in the following.
%Specifically, average node degree, average shortest path length \cite{borgwardt2005shortest}, and average clustering coefficient \cite{giatsidis2014corecluster} reflect the structural and topological properties of various datasets or domains, from which we observe that different domains exhibit unique structural characteristics, highlighting the importance of structural alignment in both multi-domain pre-training and cross-domain adaptation.

\begin{itemize}[leftmargin=*]
    \item \emph{Cora} \cite{mccallum2000automating} consists of 2,708 publications in the computing field, each categorized into one of seven classes. The citation network comprises 5,429 edges. Each publication is represented by a binary word vector indicating the presence or absence of words from a dictionary containing 1,433 unique words.
    \item \emph{Citeseer}\cite{sen2008collective} contains 3,312 computer science publications, each belonging to one of six categories, distinct from those in \textit{Cora}. The citation network consists of 4,732 edges. Each publication is represented by a binary word vector, reflecting the presence or absence of words from a dictionary of 3,703 unique words.
    \item \emph{PubMed} \cite{sen2008collective} consists of 19,717 biomedical publications related to diabetes, each classified into one of three categories. The citation network includes 44,338 edges. Each publication is represented by a TF/IDF-weighted word vector, indicating the presence of 500 unique words from the dictionary.
    \item \emph{Photo} \cite{shchur2018pitfalls} contains 7,487 products related to photography, each assigned to one of eight categories. The co-purchase network comprises 119,043 edges, representing products frequently bought together. Each product is described by a feature vector derived from its metadata and reviews, and is labeled according to its category.
    \item \emph{Computers} \cite{shchur2018pitfalls} includes 13,752 computer-related products, divided into ten categories. The co-purchase network consists of 245,861 edges, representing products that are frequently bought together. Each product is characterized by a feature vector generated from its metadata and reviews and is labeled according to its respective category.
    \item \emph{Facebook} \cite{rozemberczki2021multi} represents a page-to-page Web graph of verified Facebook pages. The nodes correspond to official Facebook pages, and the edges indicate mutual ``likes'' between these pages. Node features are derived from the descriptions provided by the page owners that outline the purpose of their pages.
    \item \emph{LastFM} \cite{rozemberczki2020characteristic} represents a social network of LastFM users, collected via the public API in March 2020. The nodes correspond to LastFM users from various Asian countries, and the edges represent mutual follower relationships. The node features are extracted based on the artists that users have liked. The associated task for this graph is multinomial node classification, where the objective is to predict each user's location, derived from the country field in their profile.
\end{itemize}

\subsection{Further Descriptions of Baselines} \label{app.baselines}
In this section, we provide additional details about the baselines used in our experiments.

\vspace{1mm}
\noindent (1) End-to-end GNNs:
\begin{itemize}[leftmargin=*]
    \item \textbf{GCN} \cite{kipf2016semi}: GCN employs a mean-pooling approach for neighborhood aggregation, enabling the integration of information from adjacent nodes.
    \item \textbf{GAT} \cite{velivckovic2017graph}: GAT relies on neighborhood aggregation for node representation learning, but distinguishes itself by assigning varying attention weights to neighbors, thus adjusting their influence on the aggregation process.
\end{itemize}

\noindent (2) Graph pre-training models:
\begin{itemize}[leftmargin=*]
    \item \textbf{DGI} \cite{velivckovic2017graph}: DGI is a self-supervised pre-training approach. It is based maximizing mutual information (MI), with the goal of strengthening the MI between local node representations and their global context.
    \item \textbf{InfoGraph} \cite{sun2019infograph}: Building on DGI, InfoGraph focuses on graph-level tasks, aiming to align node and graph embeddings by maximizing the similarity between them.
    \item \textbf{GraphCL} \cite{you2020graph}: GraphCL applies various graph augmentations for self-supervised learning, leveraging structural patterns within graphs. Its main objective is to improve the similarity across different augmentations during pre-training.
    \item \textbf{GPPT} \cite{sun2022gppt}: GPPT pre-trains a GNN model via link prediction task. Its downstream prompt module is specifically designed for node classification, unifying it with the pre-training link prediction task.
    \item \textbf{GPF} \cite{fang2022universal}: GPF serves as a universal prompt-based tuning approach for pre-trained graph models. It adapts the input graph's feature space to simulate the behavior of various prompting functions.
    \item \textbf{GraphPrompt} \cite{liu2023graphprompt}: GraphPrompt utilizes subgraph similarity calculations as a unified framework to bridge the gap between pre-training and downstream tasks, supporting both node and graph classification. During downstream adaptation, a learnable prompt is tuned to incorporate task-specific knowledge.
\end{itemize}

\noindent (3) Graph cross-domain models:
\begin{itemize}[leftmargin=*]
    \item Hassani \cite{hassani2022cross}: Hassani proposes an attention-based graph encoder that leverages both contextual and topological views to capture task-specific information for quick adaptation, as well as task-independent knowledge for efficient transfer across domains.
\end{itemize}

% \noindent (4) \textbf{Graph Prompt Models}
% \begin{itemize}
%     \item \textbf{GPPT} \cite{sun2022gppt}: GPPT pre-trains a GNN model via link prediction task. Its downstream prompt module is specifically designed for node classification, unifying it with the pre-training link prediction task.
%     \item \textbf{GPF} \cite{fang2022universal}: GPF serves as a universal prompt-based tuning approach for pre-trained graph models. It adapts the input graph's feature space to simulate the behavior of various prompting functions.
%     \item \textbf{GraphPrompt} \cite{liu2023graphprompt}: GraphPrompt utilizes subgraph similarity calculations as a unified framework to bridge the gap between pre-training and downstream tasks, supporting both node and graph classification. During downstream adaptation, a learnable prompt is tuned to incorporate task-specific knowledge.
% \end{itemize}

\noindent (4) Multi-domain graph pre-training models:
\begin{itemize}[leftmargin=*]
    \item \textbf{GCOPE} \cite{zhao2024all}: GCOPE propose a multi-domain pre-training strategy that integrates graph datasets from various domains using domain-specific interconnecting virtual nodes, which link nodes within the same domain. The main objective is to enhance downstream performance by harnessing knowledge from multiple source domains.
\end{itemize}

\subsection{Implementation Details} \label{app.parameters}

% \stitle{Environment.}
% \label{app.general-setting}
% \textbf{Optimizer:} 
% The environment in which we ran all experiments is listed below.
% \begin{itemize}
%    \item Operating system: Ubuntu 22.04.2,
%    \item CPU information: AMD EPYC 7742 64-Core Processor,
%    \item GPU information: NVIDIA GeForce RTX 3090 (24 GB).
% \end{itemize}

We outline key settings for the baselines and \model. 

\stitle{Baseline settings.}
We utilized the official codes for all open-source baselines. Each model was tuned based on the settings recommended in their respective work to achieve optimal performance.

For the baseline GCN \cite{kipf2016semi}, we employ a 3-layer architecture, and set the hidden dimensions to 256. 
For GAT \cite{velivckovic2017graph}, we employ a 2-layer architecture and set the hidden dimension to 64. Additionally, we apply 8 attention heads in the first GAT layer.

For DGI \cite{velivckovic2017graph}, we utilize a 1-layer GCN as the backbone and set the hidden dimensions to 256. Additionally, we employ prelu as the activation function.
For InfoGraph \cite{sun2019infograph}, a 3-layer GCN is used as the backbone, with its hidden dimensions set to 256.
For GraphCL \cite{you2020graph}, a 1-layer GCN is also employed as its backbone, with the hidden dimensions set to 256. Specifically, we select edge dropping as the augmentations, with a default augmentation ratio of 0.2.
For GPPT \cite{sun2022gppt}, we utilize a 2-layer GraphSAGE as its backbone, setting the hidden dimensions to 256. We employ a mean aggregator for the aggregation in the backbone.
For GraphPrompt \cite{liu2023graphprompt}, a 3-layer GCN is used as the backbone for all datasets, with the hidden dimensions set to 256.
For GPF \cite{fang2022universal}, we employ a 5-layer GCN as the backbone for all datasets, following the recommended settings. The hidden dimensions are set to 256.

For Hassani \cite{hassani2022cross}, a 3-layer GCN is used as the backbone for all datasets, with the hidden dimensions set to 256.

For GCOPE \cite{zhao2024all}, we employ a 2-layer GCN as the backbone and set the hidden dimensions to 100. Downstream adaptation is achieved through fine-tuning, as it is reported to yield the best performance in their literature.

For all baselines except GCOPE, we set the unified feature dimensions to 50, matching our \model. For GCOPE, we adhere to the recommended settings and set it to 100.

\stitle{\model\ settings.}
For our proposed \model, we utilize a 3-layer GCN as the backbone for all datasets, with the hidden dimensions set to 256. We set the unified feature dimensions to 50.

\subsection{Details about Heterophilic Datasets}\label{app.hetero}
To evaluate the robustness of \model\ across graphs with varying homophily ratios, we conducted experiments on both homophilic and heterophilic datasets in Sect.~\ref{sec.hetero}. Details of the heterophilic datasets are introduced as follows.
(1) \emph{Chameleon} \cite{rozemberczki2021multi} is a Wikipedia-based network containing 2,277 pages, categorized into five groups based on their average monthly traffic. This dataset forms a network with 36,101 edges, and the node features are derived from key nouns extracted from the Wikipedia content. %The homophily ratio is 0.23.
(2) \emph{Cornell} \cite{pei2020geom} is another webpage network consisting of 183 nodes, where each node represents a webpage, and 295 edges denoting hyperlinks between them. The node features are derived from a bag-of-words representation of the webpages. These pages are manually classified into five categories: student, project, course, staff, and faculty. %The homophily ratio is 0.22.
(3) \emph{Squirrel} \cite{rozemberczki2021multi} consists of 5,201 Wikipedia pages discussing specific topics. The pages are divided into five categories based on their average monthly traffic. This dataset forms a page-to-page network with 217,073 edges, and the node features are derived from various informative nouns present in the Wikipedia content. %The homophily ratio is 0.30.

\begin{figure}[t]
\centering 
\includegraphics[width=1\linewidth]{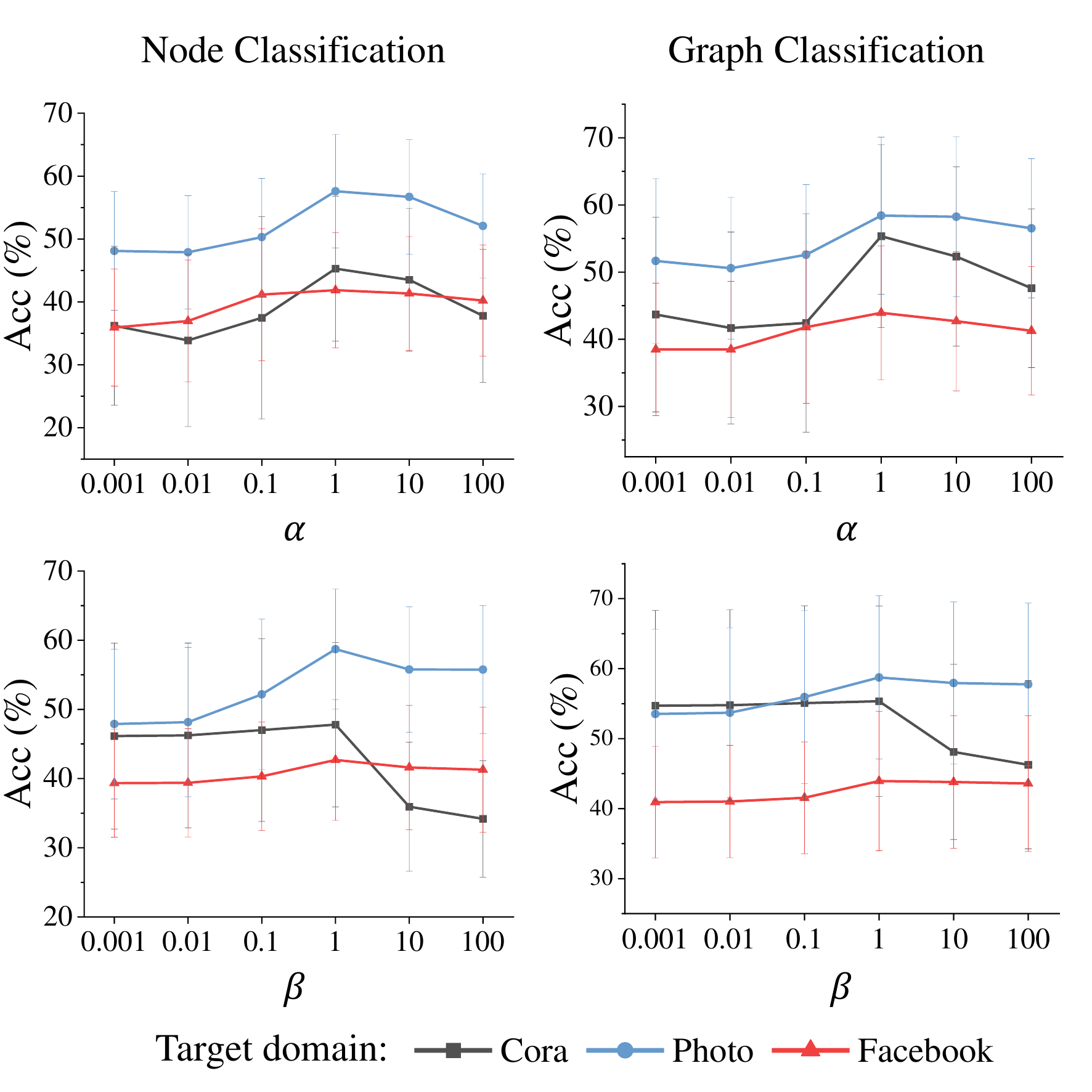}%
 \vspace{-4mm}%
\caption{Sensitivity study of $\alpha$ and $\beta$.}
\label{fig.hyperpara}
\end{figure}

\subsection{Hyperparameter Sensitivity}\label{app.hyperpara}
We investigate the impact of hyperparameters, $\alpha$ and $\beta$, in \model. $\alpha$ governs the fusion of feature and structure alignment, as well as their adaptation, in Eqs.~\eqref{eq.source-fusion} and~\eqref{eq.down-fusion}, whereas $\beta$ controls the aggregation of holistic and domain-specific adaptation in Eq.~\eqref{eq.structure-emb}. We vary $\alpha$ and $\beta$ and present 1-shot node and graph classification results on three target domain, \textit{Cora}, \textit{Photo} and \textit{Facebook}, in Fig.~\ref{fig.hyperpara}, with error bars denoting the standard deviation. 

We observe that increasing $\alpha$ from lower values initially enhances performance as structure alignment and adaptation are emphasized. However, after reaching a peak ($\alpha=1$), accuracy begins to decline as $\alpha$ grows further, implying that both feature and structure alignment are essential. 
%, further demonstrating the importance of our structural alignment design.
Moreover, $\beta$ exhibit a trend similar to that of $\alpha$, demonstrating that incorporating both holistic and domain-specific knowledge is vital for cross-domain adaptation.
Based on the above observations, we set $\alpha=1$ in our experiments, indicating a balance between the feature and structure counterparts, and $\beta=1$, indicating a balance between holistic and specific prompts, both of which show robust empirical performance.

\subsection{Data Ethics Statement}
To evaluate the efficacy of \model, we conducted experiments with only publicly available
datasets, including Cora\footnote{\url{https://github.com/shchur/gnn-benchmark/raw/master/data/npz/cora.npz}}, Citeseer\footnote{\url{https://github.com/shchur/gnn-benchmark/raw/master/data/npz/citeseer.npz}}, Pubmed\footnote{\url{https://github.com/shchur/gnn-benchmark/raw/master/data/npz/pubmed.npz}}, Photo\footnote{\url{https://github.com/shchur/gnn-benchmark/raw/master/data/npz/amazon_electronics_photo.npz}},  Computers\footnote{\url{https://github.com/shchur/gnn-benchmark/raw/master/data/npz/amazon_electronics_computers.npz}}, 
Facebook\footnote{\url{https://graphmining.ai/datasets/ptg/facebook.npz}},
LastFM\footnote{\url{https://graphmining.ai/datasets/ptg/lastfm_asia.npz}},
Chameleon\footnote{\url{https://github.com/SitaoLuan/ACM-GNN/tree/main/new_data/chameleon}},
Cornell\footnote{\url{https://github.com/bingzhewei/geom-gcn/tree/master/new_data/cornell}}, and
Squirrel\footnote{\url{https://github.com/SitaoLuan/ACM-GNN/tree/main/new_data/squirrel}}
in accordance to their usage terms and conditions, if any.
We also confirm that no personally identifiable information was utilized, and this research did not involve any human or animal subjects.

\end{document}